\documentclass[10pt,twocolumn,letterpaper]{article}

\usepackage[pagenumbers]{cvpr}

\usepackage[accsupp]{axessibility}
\usepackage{microtype}
\usepackage[dvipsnames]{xcolor}
\usepackage{pifont}
\usepackage{multicol}
\usepackage{multirow}
\usepackage{caption}
\usepackage{paralist}

\usepackage{lipsum}

\newcommand\blfootnote[1]{%
  \begingroup
  \renewcommand\thefootnote{}\footnote{#1}%
  \addtocounter{footnote}{-1}%
  \endgroup
}

\newcommand{\cmark}{\ding{51}}%

\definecolor{onlinecolor}{HTML}{D0C2B6}
\definecolor{frozencolor}{HTML}{8693AB}
\definecolor{mypink}{HTML}{FF968D}

\definecolor{cvprblue}{rgb}{0.21,0.49,0.74}
\usepackage[pagebackref,breaklinks,colorlinks,citecolor=cvprblue]{hyperref}
\usepackage{xspace}

\newcommand{\name}{Depth Anything\xspace}

\title{Depth Anything: Unleashing the Power of Large-Scale Unlabeled Data}

\author{Lihe Yang$^1$\blfootnote{$\dag$~corresponding authors}~~~~Bingyi Kang$^{2\hspace{0.3mm}\dag}$~~~~Zilong Huang$^2$~~~~Xiaogang Xu$^{3,4}$~~~~Jiashi Feng$^2$~~~~Hengshuang Zhao$^{1\hspace{0.2mm}\ddag}$\vspace{2mm}\\
$^1$HKU\hspace{1.5cm}$^2$TikTok\hspace{1.5cm}$^3$CUHK\hspace{1.5cm}$^4$ZJU\\
{\small $\dag$ project lead\hspace{0.6cm}$\ddag$ corresponding author}\\
{\tt\small \url{https://depth-anything.github.io}}
}

\begin{document}

\twocolumn[{
\maketitle
\begin{center}
    \captionsetup{type=figure}
    \vspace{-1mm}
    \includegraphics[width=1.0\textwidth, height=3.5cm]{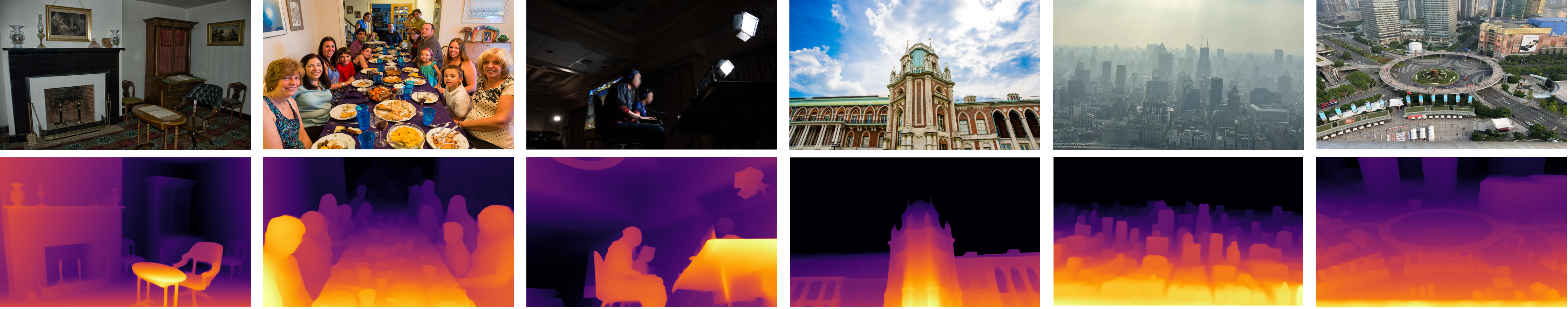}\vspace{-1mm}\\
    \captionof{figure}{Our model exhibits impressive generalization ability across extensive unseen scenes. 
    \textbf{Left two columns:} COCO \cite{coco}. \textbf{Middle two:} SA-1B \cite{sam} (a hold-out unseen set). \textbf{Right two:} photos captured by ourselves. Our model works robustly in low-light environments (1st and 3rd column), complex scenes (2nd and 5th column), foggy weather (5th column), and ultra-remote distance (5th and 6th column), \etc.}
    \label{fig:teaser}
\end{center}
}]

\begin{abstract}
\vspace{-1mm}
    \blfootnote{Work was done during an internship at TikTok.}This work presents \name\footnote{While the grammatical soundness of this name may be questionable, we treat it as a whole and pay homage to Segment Anything~\cite{sam}.}, a highly practical solution for robust monocular depth estimation. Without pursuing novel technical modules, we aim to build a simple yet powerful foundation model dealing with any images under any circumstances. 
    To this end, we scale up the dataset by designing a data engine to collect and automatically annotate large-scale unlabeled data ($\sim$62M), which significantly enlarges the data coverage and thus is able to reduce the generalization error.
    We investigate two simple yet effective strategies that make data scaling-up promising. First, a more challenging optimization target is created by leveraging data augmentation tools. It compels the model to actively seek extra visual knowledge and acquire robust representations. Second, an auxiliary supervision is developed to enforce the model to inherit rich semantic priors from pre-trained encoders. We evaluate its zero-shot capabilities extensively, including six public datasets and randomly captured photos. 
    It demonstrates impressive generalization ability (Figure~\ref{fig:teaser}).
    Further, through fine-tuning it with metric depth information from NYUv2 and KITTI, new SOTAs are set. Our better depth model also results in a better depth-conditioned ControlNet. 
    Our models are released \href{https://github.com/LiheYoung/Depth-Anything}{here}.
\end{abstract}
    
\section{Introduction}

The field of computer vision and natural language processing is currently experiencing a revolution with the emergence of ``foundation models"~\cite{bommasani2021opportunities} that demonstrate strong zero-/few-shot performance in various downstream scenarios~\cite{clip, llama}.
These successes primarily rely on large-scale training data that can effectively cover the data distribution. 
Monocular Depth Estimation (MDE), which is a fundamental problem with broad applications in robotics~\cite{wofk2019fastdepth}, autonomous driving~\cite{wang2019pseudo, you2020pseudo}, virtual reality~\cite{rasla2022relative}, \etc, also requires a foundation model to estimate depth information from a single image.
However, this has been underexplored due to the difficulty of building datasets with tens of millions of depth labels. MiDaS~\cite{midas} made a pioneering study along this direction by training an MDE model on a collection of mixed labeled datasets. Despite demonstrating a certain level of zero-shot ability, MiDaS is limited by its data coverage, thus suffering disastrous performance in some scenarios.

In this work, our goal is to \emph{build a foundation model for MDE} capable of producing high-quality depth information for any images under any circumstances. We approach this target from the perspective of dataset scaling-up. 
Traditionally, depth datasets are created mainly by acquiring depth data from sensors~\cite{kitti, nyu}, stereo matching~\cite{cityscapes}, or SfM~\cite{megadepth}, which is costly, time-consuming, or even intractable in particular situations.
We instead, for the first time, pay attention to large-scale unlabeled data. Compared with stereo images or labeled images from depth sensors, our used monocular unlabeled images exhibit three advantages:
\begin{inparaenum}[(i)]
\item (\emph{simple and cheap to acquire}) Monocular images exist almost everywhere, thus they are easy to collect, without requiring specialized devices.
\item (\emph{diverse}) Monocular images can cover a broader range of scenes, which are critical to the model generalization ability and scalability.
\item (\emph{easy to annotate}) We can simply use a pre-trained MDE model to assign depth labels for unlabeled images, which only takes a feedforward step. More than efficient, this also produces denser depth maps than LiDAR~\cite{kitti} and omits the computationally intensive stereo matching process.
\end{inparaenum}

We design a data engine to automatically generate depth annotations for unlabeled images, enabling data scaling-up to arbitrary scale. It collects 62M diverse and informative images from eight public large-scale datasets, \eg, SA-1B~\cite{sam}, Open Images~\cite{openimage}, and BDD100K~\cite{bdd100k}. We use their raw unlabeled images without any forms of labels. Then, in order to provide a reliable annotation tool for our unlabeled images, we collect 1.5M labeled images from six public datasets to train an initial MDE model. The unlabeled images are then automatically annotated and jointly learned with labeled images in a self-training manner~\cite{pseudolabel}.

Despite all the aforementioned advantages of monocular unlabeled images, it is indeed not trivial to make positive use of such large-scale unlabeled images~\cite{yalniz2019billion, zoph2020rethinking}, especially in the case of sufficient labeled images and strong pre-training models. In our preliminary attempts, directly combining labeled and pseudo labeled images failed to improve the baseline of solely using labeled images. We conjecture that, the additional knowledge acquired in such a naive self-teaching manner is rather limited. To address the dilemma, we propose to challenge the student model with a more difficult optimization target when learning the pseudo labels.
The student model is enforced to seek extra visual knowledge and learn robust representations under various strong perturbations to better handle unseen images.

Furthermore, there have been some works~\cite{chen2019towards, guizilini2020semantically} demonstrating the benefit of an auxiliary semantic segmentation task for MDE. We also follow this research line, aiming to equip our model with better high-level scene understanding capability. However, we observed when an MDE model is already powerful enough, it is hard for such an auxiliary task to bring further gains. We speculate that it is due to severe loss in semantic information when decoding an image into a discrete class space. Therefore, considering the excellent performance of DINOv2 in semantic-related tasks, we propose to maintain the rich semantic priors from it with a simple feature alignment loss. This not only enhances the MDE performance, but also yields a multi-task encoder for both middle-level and high-level perception tasks.

Our contributions are summarized as follows:

\begin{itemize}
\setlength\itemsep{0mm}
    \item We highlight the value of data scaling-up of massive, cheap, and diverse unlabeled images for MDE.

    \item We point out a key practice in jointly training large-scale labeled and unlabeled images. Instead of learning raw unlabeled images directly, we challenge the model with a harder optimization target for extra knowledge.

    \item We propose to inherit rich semantic priors from pre-trained encoders for better scene understanding, rather than using an auxiliary semantic segmentation task.
    
    \item Our model exhibits stronger zero-shot capability than MiDaS-BEiT$_{\textrm{L-512}}$~\cite{midasv31}. Further, fine-tuned with metric depth, it outperforms ZoeDepth~\cite{zoedepth} significantly.
\end{itemize}

\section{Related Work}

\textbf{Monocular depth estimation (MDE).} 
Early works~\cite{hoiem2007recovering, liu2008sift, make3d} primarily relied on handcrafted features and traditional computer vision techniques. They were limited by their reliance on explicit depth cues and struggled to handle complex scenes with occlusions and textureless regions.

Deep learning-based methods have revolutionized monocular depth estimation by effectively learning depth representations from delicately annotated datasets \cite{nyu, kitti}. Eigen \etal~\cite{eigen2014depth} first proposed a multi-scale fusion network to regress the depth.
Following this, many works consistently improve the depth estimation accuracy by carefully designing the regression task as a classification task~\cite{adabins, binsformer}, introducing more priors~\cite{li2015depth, newcrfs, gedepth, nddepth}, and better objective functions~\cite{hrwsi, vnl}, \etc. Despite the promising performance, they are hard to generalize to unseen domains.

\vspace{1.5mm}
\noindent
\textbf{Zero-shot depth estimation.} 
Our work belongs to this research line. We aim to train an MDE model with a diverse training set and thus can predict the depth for any given image. Some pioneering works~\cite{diw, redweb} explored this direction 
by collecting more training images, but their supervision is very sparse and is only enforced on limited pairs of points.

To enable effective multi-dataset joint training, a milestone work MiDaS~\cite{midas} utilizes an affine-invariant loss to ignore the potentially different depth scales and shifts across varying datasets. Thus, MiDaS provides relative depth information. Recently, some works~\cite{metric3d, zoedepth, zerodepth} take a step further to estimate the metric depth. However, in our practice, we observe such methods exhibit poorer generalization ability than MiDaS, especially its latest version~\cite{midasv31}. Besides, as demonstrated by ZoeDepth~\cite{zoedepth}, a strong relative depth estimation model can also work well in generalizable metric depth estimation by fine-tuning with metric depth information. Therefore, we still follow MiDaS in relative depth estimation, but further strengthen it by highlighting the value of large-scale monocular unlabeled images. 

\vspace{1.5mm}
\noindent
\textbf{Leveraging unlabeled data.}
This belongs to the research area of semi-supervised learning~\cite{pseudolabel, zoph2020rethinking, fixmatch}, which is popular with various applications~\cite{softteacher, unimatch}. However, existing works typically assume only limited images are available. They rarely consider the challenging but realistic scenario where there are already sufficient labeled images but also larger-scale unlabeled images. We take this challenging direction for zero-shot MDE. We demonstrate that unlabeled images can significantly enhance the data coverage and thus improve model generalization and robustness.
\section{Depth Anything}

\begin{figure*}[t]
    \centering
    \includegraphics[width=\linewidth]{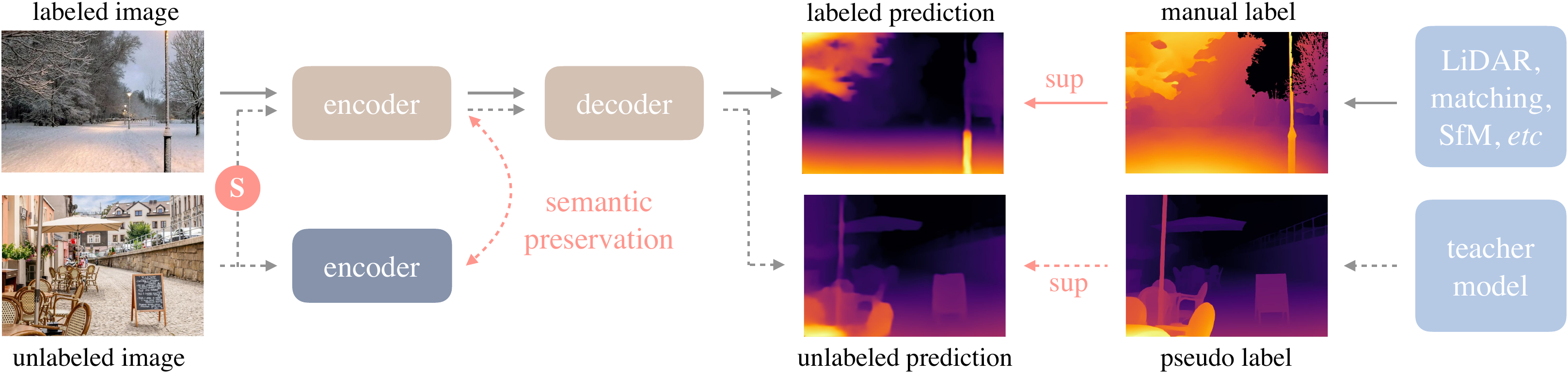}
    \caption{Our pipeline. Solid line: flow of labeled images, dotted line: unlabeled images. We especially highlight the value of large-scale unlabeled images. The \textcolor{mypink}{\textbf{S}} denotes adding strong perturbations (Section~\ref{sec:unlabeled}). To equip our depth estimation model with rich semantic priors, we enforce an auxiliary constraint between the \textcolor{onlinecolor}{\textbf{online student model}} and a \textcolor{frozencolor}{\textbf{frozen encoder}} to preserve the semantic capability (Section~\ref{sec:semantic}).}
    \vspace{-3mm}
    \label{fig:main}
\end{figure*}

Our work utilizes both labeled and unlabeled images to facilitate better monocular depth estimation (MDE). Formally, the labeled and unlabeled sets are denoted as $\mathcal{D}^l = \{(x_i, d_i)\}_{i=1}^M$ and $\mathcal{D}^u = \{u_i\}_{i=1}^N$ respectively. We aim to learn a teacher model $T$ from $\mathcal{D}^l$. Then, we utilize $T$ to assign pseudo depth labels for $\mathcal{D}^u$. Finally, we train a student model $S$ on the combination of labeled set and pseudo labeled set. A brief illustration is provided in Figure~\ref{fig:main}.

\subsection{Learning Labeled Images\label{sec:labeled}}

This process is similar to the training of MiDaS~\cite{midas, midasv31}. However, since MiDaS did not release its code, we first reproduced it. Concretely, the depth value is first transformed into the disparity space by $d = 1/t$ and then normalized to 0$\sim$1 on each depth map. To enable multi-dataset joint training, we adopt the affine-invariant loss to ignore the unknown scale and shift of each sample:
\begin{equation}
    \mathcal{L}_l = \frac{1}{HW}\sum_{i=1}^{HW}\rho(d_i^*, d_i), 
\end{equation}
where $d^*_i$ and $d_i$ are the prediction and ground truth, respectively. And $\rho$ is the affine-invariant mean absolute error loss: $\rho(d^*_i, d_i) = |\hat{d}^*_i - \hat{d}_i|$, where $\hat{d}^*_i$ and $\hat{d}_i$ are the scaled and shifted versions of the prediction $d^*_i$ and ground truth $d_i$:
\begin{equation}
    \hat{d}_i = \frac{d_i - t(d)}{s(d)},
\end{equation}
where $t(d)$ and $s(d)$ are used to align the prediction and ground truth to have zero translation and unit scale:
\begin{equation}
\label{eq:median}
    t(d) = \textrm{median}(d),\hspace{2mm} s(d) = \frac{1}{HW}\sum_{i=1}^{HW}|d_i - t(d)|.
\end{equation}

\begin{table}[t]
\small
    \centering
    \setlength\tabcolsep{1.4mm}
    \begin{tabular}{lcclr}
    \toprule
    Dataset & Indoor & Outdoor & Label & \# Images \\
    \midrule

    \multicolumn{5}{c}{Labeled Datasets} \\

    \midrule
    
    BlendedMVS \cite{blendedmvs} & \cmark & \cmark & Stereo & 115K \\
    
    DIML \cite{diml} & \cmark & \cmark & Stereo & 927K \\

    HRWSI \cite{hrwsi} & \cmark & \cmark & Stereo & 20K \\

    IRS \cite{irs} & \cmark & & Stereo & 103K \\
    
    MegaDepth \cite{megadepth} & & \cmark & SfM & 128K \\

    TartanAir \cite{tartanair} & \cmark & \cmark & Stereo & 306K \\
    
    \midrule
    
    \multicolumn{5}{c}{Unlabeled Datasets} \\

    \midrule

    BDD100K \cite{bdd100k} & & \cmark & None & 8.2M \\

    Google Landmarks \cite{googlelandmarks} & & \cmark & None & 4.1M \\

    ImageNet-21K \cite{imagenet} & \cmark & \cmark & None & 13.1M \\
    
    LSUN \cite{lsun} & \cmark & & None & 9.8M \\
    
    Objects365 \cite{objects365} & \cmark & \cmark & None & 1.7M \\
    
    Open Images V7 \cite{openimage} & \cmark & \cmark & None & 7.8M \\

    Places365 \cite{places365} & \cmark & \cmark & None & 6.5M \\
    
    SA-1B \cite{sam} & \cmark & \cmark & None & 11.1M \\
    
    \bottomrule
    \end{tabular}
    \caption{In total, our Depth Anything is trained on 1.5M labeled images and \textbf{62M unlabeled images} jointly.}
    \label{tab:dataset_info}
\end{table}

To obtain a robust monocular depth estimation model, we collect 1.5M labeled images from 6 public datasets. Details of these datasets are listed in Table~\ref{tab:dataset_info}. We use fewer labeled datasets than MiDaS v3.1~\cite{midasv31} (12 training datasets), because 1) we do not use NYUv2~\cite{nyu} and KITTI~\cite{kitti} datasets to ensure zero-shot evaluation on them, 2) some datasets are not available (anymore), \eg, Movies~\cite{midas} and WSVD \cite{wsvd}, and 3) some datasets exhibit poor quality, \eg, RedWeb (also low resolution)~\cite{redweb}. Despite using fewer labeled images, our easy-to-acquire and diverse unlabeled images will comprehend the data coverage and greatly enhance the model generalization ability and robustness.

Furthermore, to strengthen the teacher model $T$ learned from these labeled images, we adopt the DINOv2~\cite{dinov2} pre-trained weights to initialize our encoder. In practice, we apply a pre-trained semantic segmentation model~\cite{segformer} to detect the \texttt{sky} region, and set its disparity value as 0 (farthest).

\subsection{Unleashing the Power of Unlabeled Images\label{sec:unlabeled}}

This is the main point of our work. Distinguished from prior works that laboriously construct diverse labeled datasets, we highlight the value of unlabeled images in enhancing the data coverage. Nowadays, we can practically build a diverse and large-scale unlabeled set from the Internet or public datasets of various tasks. Also, we can effortlessly obtain the dense depth map of monocular unlabeled images simply by forwarding them to a pre-trained well-performed MDE model. This is much more convenient and efficient than performing stereo matching or SfM reconstruction for stereo images or videos. We select eight large-scale public datasets as our unlabeled sources for their diverse scenes. They contain more than 62M images in total. The details are provided in the bottom half of Table~\ref{tab:dataset_info}.

Technically, given the previously obtained MDE teacher model $T$, we make predictions on the unlabeled set $\mathcal{D}^u$ to obtain a pseudo labeled set $\hat{\mathcal{D}}^u$:
\begin{equation}
    \hat{\mathcal{D}}^u = \{(u_i, T(u_i)) | u_i \in \mathcal{D}^u\}_{i=1}^N.
\end{equation}

With the combination set $\mathcal{D}^l \cup \hat{\mathcal{D}^u}$ of labeled images and pseudo labeled images, we train a student model $S$ on it. Following prior works~\cite{st++}, instead of fine-tuning $S$ from $T$, we re-initialize $S$ for better performance.

Unfortunately, in our pilot studies, we failed to gain improvements with such a self-training pipeline, which indeed contradicts the observations when there are only a few labeled images~\cite{fixmatch}. We conjecture that, with already sufficient labeled images in our case, the extra knowledge acquired from additional unlabeled images is rather limited. Especially considering the teacher and student share the same pre-training and architecture, they tend to make similar correct or false predictions on the unlabeled set $\mathcal{D}^u$, even without the explicit self-training procedure.

To address the dilemma, we propose to challenge the student with a more difficult optimization target for additional visual knowledge on unlabeled images. We inject strong perturbations to unlabeled images during training. It compels our student model to actively seek extra visual knowledge and acquire invariant representations from these unlabeled images. These advantages help our model deal with the open world more robustly. We introduce two forms of perturbations: one is strong color distortions, including color jittering and Gaussian blurring, and the other is strong spatial distortion, which is CutMix~\cite{cutmix}. Despite the simplicity, the two modifications make our large-scale unlabeled images significantly improve the baseline of labeled images.

We provide more details about CutMix. It was originally proposed for image classification, and is rarely explored in monocular depth estimation. We first interpolate a random pair of unlabeled images $u_a$ and $u_b$ spatially:
\begin{equation}
    u_{ab} = u_a \odot M + u_b \odot (1 - M),
\end{equation}
where $M$ is a binary mask with a rectangle region set as 1.

The unlabeled loss $\mathcal{L}_u$ is obtained by first computing affine-invariant losses in valid regions defined by $M$ and $1-M$, respectively:
\begin{align}
    &\mathcal{L}^M_u = \rho\big(S(u_{ab}) \odot M, \,T(u_a) \odot M\big),\\
    &\mathcal{L}^{1-M}_u = \rho\big(S(u_{ab}) \odot (1-M), T(u_b) \odot (1-M)\big),
\end{align}
where we omit the $\sum$ and pixel subscript $i$ for simplicity.
Then we aggregate the two losses via weighted averaging:
\begin{equation}
    \mathcal{L}_u = \frac{\sum M}{HW}\mathcal{L}^M_u + \frac{\sum (1-M)}{HW}\mathcal{L}^{1-M}_u.
\end{equation}

We use CutMix with 50\% probability. The unlabeled images for CutMix are already strongly distorted in color, but the unlabeled images fed into the teacher model $T$ for pseudo labeling are clean, without any distortions. 

\begin{table*}[t]
\small
    \centering
    \setlength\tabcolsep{1.5mm}
    \begin{tabular}{lccccccccccccc}
    \toprule
    \multirow{2}{*}{Method} & \multirow{2}{*}{Encoder} & \multicolumn{2}{c}{KITTI~\cite{kitti}} & \multicolumn{2}{c}{NYUv2~\cite{nyu}} & \multicolumn{2}{c}{Sintel~\cite{sintel}} & \multicolumn{2}{c}{DDAD~\cite{ddad}} & \multicolumn{2}{c}{ETH3D~\cite{eth3d}} & \multicolumn{2}{c}{DIODE~\cite{diode}} \\

    \cmidrule(lr){3-4}\cmidrule(lr){5-6}\cmidrule(lr){7-8}\cmidrule(lr){9-10}\cmidrule(lr){11-12}\cmidrule(lr){13-14}
    
    ~ & ~ & AbsRel & $\delta_1$ & AbsRel & $\delta_1$ & AbsRel & $\delta_1$ & AbsRel & $\delta_1$ & AbsRel & $\delta_1$ & AbsRel & $\delta_1$ \\
    
    \midrule

    MiDaS v3.1~\cite{midasv31} & ViT-L & 0.127 & 0.850 & 0.048 & \underline{0.980} & 0.587 & 0.699 & 0.251 & 0.766 & 0.139 & 0.867 & 0.075 & 0.942 \\

    \midrule
    
    \multirow{3}{*}{\textbf{Depth Anything}} & ViT-S & 0.080 & 0.936 & 0.053 & 0.972 & 0.464 & 0.739 & 0.247 & 0.768 & 0.127 & \textbf{0.885} &  0.076 & 0.939 \\

    ~ & ViT-B & \underline{0.080} & \underline{0.939} & \underline{0.046} & 0.979 & \textbf{0.432} & \underline{0.756} & \underline{0.232} & \underline{0.786} & \textbf{0.126} & \underline{0.884} & \underline{0.069} & \underline{0.946} \\

    ~ & ViT-L & \textbf{0.076} & \textbf{0.947} & \textbf{0.043} & \textbf{0.981} & \underline{0.458} & \textbf{0.760} & \textbf{0.230} & \textbf{0.789} & \underline{0.127} & 0.882 & \textbf{0.066} & \textbf{0.952} \\
    
    \bottomrule
    \end{tabular}
    \caption{\textbf{Zero-shot relative} depth estimation. \textbf{Better:} AbsRel $\downarrow$ , $\delta_1$ $\uparrow$. We compare with the best model from MiDaS v3.1. Note that MiDaS \textbf{\emph{does not}} strictly follow the zero-shot evaluation on KITTI and NYUv2, because it uses their training images. We provide three model scales for different purposes, based on ViT-S (24.8M), ViT-B (97.5M), and ViT-L (335.3M), respectively. \textbf{Best}, \underline{second best} results.}
    \label{tab:zeroshot_mde}
\end{table*}

\subsection{Semantic-Assisted Perception\label{sec:semantic}}

There exist some works~\cite{chen2019towards, guizilini2020semantically, klingner2020self, mtformer} improving depth estimation with an auxiliary semantic segmentation task. We believe that arming our depth estimation model with such high-level semantic-related information is beneficial. Besides, in our specific context of leveraging unlabeled images, these auxiliary supervision signals from other tasks can also combat the potential noise in our pseudo depth label.

Therefore, we made an initial attempt by carefully assigning semantic segmentation labels to our unlabeled images with a combination of RAM~\cite{ram} + GroundingDINO~\cite{groundingdino} + HQ-SAM~\cite{hqsam} models. After post-processing, this yields a class space containing 4K classes. In the joint-training stage, the model is enforced to produce both depth and segmentation predictions with a shared encoder and two individual decoders. Unfortunately, after trial and error, we still could not boost the performance of the original MDE model. We speculated that, decoding an image into a discrete class space indeed loses too much semantic information. The limited information in these semantic masks is hard to further boost our depth model, especially when our depth model has established very competitive results.

Therefore, we aim to seek more informative semantic signals to serve as auxiliary supervision for our depth estimation task. We are greatly astonished by the strong performance of DINOv2 models~\cite{dinov2} in semantic-related tasks, \eg, image retrieval and semantic segmentation, even with frozen weights without any fine-tuning. Motivated by these clues, we propose to transfer its strong semantic capability to our depth model with an auxiliary feature alignment loss. The feature space is high-dimensional and continuous, thus containing richer semantic information than discrete masks. The feature alignment loss is formulated as:
\begin{equation}
    \mathcal{L}_{feat} = 1 - \frac{1}{HW}\sum_{i=1}^{HW}\cos(f_i, f'_i),
\end{equation}
where $\cos(\cdot, \cdot)$ measures the cosine similarity between two feature vectors. $f$ is the feature extracted by the depth model $S$, while $f'$ is the feature from a frozen DINOv2 encoder. We do not follow some works~\cite{byol} to project the online feature $f$ into a new space for alignment, because a randomly initialized projector makes the large alignment loss dominate the overall loss in the early stage.

Another key point in feature alignment is that, semantic encoders like DINOv2 tend to produce similar features for different parts of an object, \eg, car front and rear. In depth estimation, however, different parts or even pixels within the same part, can be of varying depth. Thus, it is not beneficial to \emph{exhaustively} enforce our depth model to produce exactly the same features as the frozen encoder. 

To solve this issue, we set a tolerance margin $\alpha$ for the feature alignment. If the cosine similarity of $f_i$ and $f'_i$ has surpassed $\alpha$, this pixel will not be considered in our $\mathcal{L}_{feat}$. This allows our method to enjoy both the semantic-aware representation from DINOv2 and the part-level discriminative representation from depth supervision. As a side effect, our produced encoder not only performs well in downstream MDE datasets, but also achieves strong results in the semantic segmentation task. It also indicates the potential of our encoder to serve as a universal multi-task encoder for both middle-level and high-level perception tasks.

Finally, our overall loss is an average combination of the three losses $\mathcal{L}_l$, $\mathcal{L}_u$, and $\mathcal{L}_{feat}$.

\section{Experiment}

\subsection{Implementation Details\label{sec:detail}}

We adopt the DINOv2 encoder~\cite{dinov2} for feature extraction. Following MiDaS~\cite{midas, midasv31}, we use the DPT~\cite{dpt} decoder for depth regression. All labeled datasets are simply combined together without re-sampling. In the first stage, we train a teacher model on labeled images for 20 epochs. In the second stage of joint training, we train a student model to sweep across all unlabeled images for one time. The unlabeled images are annotated by a best-performed teacher model with a ViT-L encoder. The ratio of labeled and unlabeled images is set as 1:2 in each batch. In both stages, the base learning rate of the pre-trained encoder is set as 5e-6, while the randomly initialized decoder uses a 10$\times$ larger learning rate. We use the AdamW optimizer and decay the learning rate with a linear schedule. We only apply horizontal flipping as our data augmentation for labeled images. The tolerance margin $\alpha$ for feature alignment loss is set as 0.85. For more details, please refer to our appendix.

\subsection{Zero-Shot Relative Depth Estimation\label{sec:zero_depth}}

As aforementioned, this work aims to provide accurate depth estimation for any image. Therefore, we comprehensively validate the zero-shot depth estimation capability of our Depth Anything model on six representative unseen datasets: KITTI~\cite{kitti}, NYUv2~\cite{nyu}, Sintel~\cite{sintel}, DDAD~\cite{ddad}, ETH3D~\cite{eth3d}, and DIODE~\cite{diode}. We compare with the best DPT-BEiT$_{\textrm{L-512}}$ model from the latest MiDaS v3.1~\cite{midasv31}, which uses more labeled images than us. As shown in Table~\ref{tab:zeroshot_mde}, both with a ViT-L encoder, our Depth Anything surpasses the strongest MiDaS model tremendously across extensive scenes in terms of both the AbsRel (absolute relative error: $|d^* - d| / d$) and $\delta_1$ (percentage of $\max(d^*/d, d/d^*) < 1.25$) metrics. For example, when tested on the well-known autonomous driving dataset DDAD~\cite{ddad}, we improve the AbsRel ($\downarrow$) from 0.251 $\rightarrow$ 0.230 and improve the $\delta_1$ ($\uparrow$) from 0.766 $\rightarrow$ 0.789. 

Besides, our ViT-B model is already clearly superior to the MiDaS based on a much larger ViT-L. Moreover, our ViT-S model, whose scale is less than 1/10 of the MiDaS model, even outperforms MiDaS on several unseen datasets, including Sintel, DDAD, and ETH3D. The performance advantage of these small-scale models demonstrates their great potential in computationally-constrained scenarios.

It is also worth noting that, on the most widely used MDE benchmarks KITTI and NYUv2, although MiDaS v3.1 uses the corresponding training images (\emph{not zero-shot anymore}), our Depth Anything is still evidently superior to it \emph{without training with any KITTI or NYUv2 images}, \eg, 0.127 \emph{vs.} 0.076 in AbsRel and 0.850 \emph{vs.} 0.947 in $\delta_1$ on KITTI.

\subsection{Fine-tuned to \textbf{\emph{Metric}} Depth Estimation\label{sec:finetune_depth}}

Apart from the impressive performance in zero-shot relative depth estimation, we further examine our Depth Anything model as a promising weight initialization for downstream \emph{metric} depth estimation. We initialize the encoder of downstream MDE models with our pre-trained encoder parameters and leave the decoder randomly initialized. The model is fine-tuned with correponding metric depth information. In this part, we use our ViT-L encoder for fine-tuning.

We examine two representative scenarios: 1) \emph{in-domain} metric depth estimation, where the model is trained and evaluated on the same domain (Section~\ref{sec:indomain_metric}), and 2) \emph{zero-shot} metric depth estimation, where the model is trained on one domain, \eg, NYUv2~\cite{nyu}, but evaluated in different domains, \eg, SUN RGB-D~\cite{sunrgbd} (Section~\ref{sec:zeroshot_metric}).

\subsubsection{In-Domain Metric Depth Estimation\label{sec:indomain_metric}}

As shown in Table~\ref{tab:nyu_finetune} of NYUv2~\cite{nyu}, our model outperforms the previous best method VPD~\cite{vpd} remarkably, improving the $\delta_1$ ($\uparrow$) from 0.964 $\rightarrow$ 0.984 and AbsRel ($\downarrow$) from 0.069 to 0.056. Similar improvements can be observed in Table~\ref{tab:kitti_finetune} of the KITTI dataset~\cite{kitti}. We improve the $\delta_1$ ($\uparrow$) on KITTI from 0.978  $\rightarrow$ 0.982. It is worth noting that we adopt the ZoeDepth framework for this scenario with a relatively basic depth model, and we believe our results can be further enhanced if equipped with more advanced architectures.

\begin{table}[t]
\footnotesize
    \centering
    \setlength\tabcolsep{1.75mm}
    \renewcommand{\arraystretch}{1.05}
    \begin{tabular}{rcccccc}
    \toprule
    \multirow{2}{*}{Method} & \multicolumn{3}{c}{\emph{Higher is better} $\uparrow$} & \multicolumn{3}{c}{\emph{Lower is better} $\downarrow$} \\

    \cmidrule(lr){2-4}\cmidrule(lr){5-7}
    
    ~ & \textcolor{cvprblue}{$\delta_1$} & $\delta_2$ & $\delta_3$ & \textcolor{cvprblue}{AbsRel} & \textcolor{cvprblue}{RMSE} & log10 \\
    
    \midrule 
    
    AdaBins~\cite{adabins} & 0.903 & 0.984 & 0.997 & 0.103 & 0.364 & 0.044 \\
    
    DPT~\cite{dpt} & 0.904 & 0.988 & 0.998 & 0.110 & 0.357 & 0.045 \\

    P3Depth~\cite{p3depth} & 0.898 & 0.981 & 0.996 & 0.104 & 0.356 & 0.043 \\
    
    SwinV2-L~\cite{swinv2} & 0.949 & 0.994 & 0.999 & 0.083 & 0.287 & 0.035 \\
    
    AiT~\cite{ait} & 0.954 & 0.994 & 0.999 & 0.076 & 0.275 & 0.033 \\

    VPD~\cite{vpd} & \underline{0.964} & \underline{0.995} & \underline{0.999} & \underline{0.069} & \underline{0.254} & \underline{0.030} \\

    ZoeDepth$^*$~\cite{zoedepth} & 0.951 & 0.994 & 0.999 & 0.077 & 0.282 & 0.033 \\
        
    \midrule
    
    \textbf{Ours} & \textbf{0.984} & \textbf{0.998} & \textbf{1.000} & \textbf{0.056} & \textbf{0.206} & \textbf{0.024} \\
    
    \bottomrule
    \end{tabular}
    \caption{\textbf{Fine-tuning and evaluating on NYUv2}~\cite{nyu} with our pre-trained MDE encoder. We highlight \textbf{best}, \underline{second best} results, as well as \textcolor{cvprblue}{\textbf{most discriminative metrics}}. $*$: Reproduced by us.}
    \label{tab:nyu_finetune}
\end{table}

\begin{table}[t]
\footnotesize
    \centering
    \setlength\tabcolsep{1.35mm}
    \renewcommand{\arraystretch}{1.05}
    \begin{tabular}{rcccccc}
    \toprule
    \multirow{2}{*}{Method} & \multicolumn{3}{c}{\emph{Higher is better} $\uparrow$} & \multicolumn{3}{c}{\emph{Lower is better} $\downarrow$} \\

    \cmidrule(lr){2-4}\cmidrule(lr){5-7}
    
    ~ & \textcolor{cvprblue}{$\delta_1$} & $\delta_2$ & $\delta_3$ & \textcolor{cvprblue}{AbsRel} & \textcolor{cvprblue}{RMSE} & RMSE log \\

    \midrule

    AdaBins~\cite{adabins} & 0.964 & 0.995 & 0.999 & 0.058 & 2.360 & 0.088 \\
    
    DPT~\cite{dpt} & 0.959 & 0.995 & 0.999 & 0.062 & 2.573 & 0.092 \\

    P3Depth~\cite{p3depth} & 0.953 & 0.993 & 0.998 & 0.071 & 2.842 & 0.103 \\
    
    NeWCRFs~\cite{newcrfs} & 0.974 & 0.997 & 0.999 & 0.052 & 2.129 & 0.079 \\
    
    SwinV2-L~\cite{swinv2} & 0.977 & 0.998 & \underline{1.000} & 0.050 & \underline{1.966} & \underline{0.075} \\

    NDDepth~\cite{nddepth} & \underline{0.978} & \underline{0.998} & 0.999 & 0.050 & 2.025 & 0.075 \\

    GEDepth~\cite{gedepth} & 0.976 & 0.997 & 0.999 & \underline{0.048} & 2.044 & 0.076 \\   

    ZoeDepth$^*$~\cite{zoedepth} & 0.971 & 0.996 & 0.999 & 0.054 & 2.281 & 0.082 \\
    
    \midrule
    
    \textbf{Ours} & \textbf{0.982} & \textbf{0.998} & \textbf{1.000} & \textbf{0.046} & \textbf{1.896} & \textbf{0.069} \\

    \bottomrule
    \end{tabular}
    \caption{\textbf{Fine-tuning and evaluating on KITTI}~\cite{kitti} with our pre-trained MDE encoder. $*$: Reproduced by us.}
    \label{tab:kitti_finetune}
\end{table}

\begin{table*}[t]
\small
    \centering
    \setlength\tabcolsep{2.55mm}
    \begin{tabular}{lcccccccccc}
    \toprule
    \multirow{2}{*}{Method} & \multicolumn{2}{c}{SUN RGB-D~\cite{sunrgbd}} & \multicolumn{2}{c}{iBims-1~\cite{ibims}} & \multicolumn{2}{c}{HyperSim~\cite{hypersim}} & \multicolumn{2}{c}{Virtual KITTI 2~\cite{vkitti}} & \multicolumn{2}{c}{DIODE Outdoor~\cite{diode}} \\

    \cmidrule(lr){2-3}\cmidrule(lr){4-5}\cmidrule(lr){6-7}\cmidrule(lr){8-9}\cmidrule(lr){10-11}
    
    ~ & AbsRel ($\downarrow$) & $\delta_1$ ($\uparrow$) & AbsRel & $\delta_1$ & AbsRel & $\delta_1$ & AbsRel & $\delta_1$ & AbsRel & $\delta_1$ \\
    
    \midrule

    ZoeDepth~\cite{zoedepth} & 0.520 & 0.545 & 0.169 & 0.656 & 0.407 & 0.302 & 0.106 & 0.844 & 0.814 & 0.237 \\
    
    \textbf{Depth Anything} & \textbf{0.500} & \textbf{0.660} & \textbf{0.150} & \textbf{0.714} & \textbf{0.363} & \textbf{0.361} & \textbf{0.085} & \textbf{0.913} & \textbf{0.794} & \textbf{0.288} \\
    
    \bottomrule
    \end{tabular}
    \caption{\textbf{Zero-shot metric} depth estimation. The first three test sets in the header are indoor scenes, while the last two are outdoor scenes. Following ZoeDepth, we use the model trained on NYUv2 for indoor generalization, while use the model trained on KITTI for outdoor evaluation. For fair comparisons, we report the ZoeDepth results reproduced in our environment.}
    \label{tab:zeroshot_metric_mde}
\end{table*}

\subsubsection{Zero-Shot Metric Depth Estimation\label{sec:zeroshot_metric}}

We follow ZoeDepth~\cite{zoedepth} to conduct zero-shot metric depth estimation. ZoeDepth fine-tunes the MiDaS pre-trained encoder with metric depth information from NYUv2~\cite{nyu} (for indoor scenes) or KITTI~\cite{kitti} (for outdoor scenes). Therefore, we simply replace the MiDaS encoder with our better Depth Anything encoder, leaving other components unchanged. As shown in Table~\ref{tab:zeroshot_metric_mde}, 
across a wide range of unseen datasets of indoor and outdoor scenes, our Depth Anything results in a better metric depth estimation model than the original ZoeDepth based on MiDaS.

\begin{table*}[t]
\small
    \centering
    \setlength\tabcolsep{1.15mm}
    \begin{tabular}{lcccccccccccccc}
    \toprule
    \multirow{2}{*}{Training set} & \multicolumn{2}{c}{KITTI~\cite{kitti}} & \multicolumn{2}{c}{NYUv2~\cite{nyu}} & \multicolumn{2}{c}{Sintel~\cite{sintel}} & \multicolumn{2}{c}{DDAD~\cite{ddad}} & \multicolumn{2}{c}{ETH3D~\cite{eth3d}} & \multicolumn{2}{c}{DIODE~\cite{diode}} & \multicolumn{2}{c}{\textbf{Mean}} \\

    \cmidrule(lr){2-3}\cmidrule(lr){4-5}\cmidrule(lr){6-7}\cmidrule(lr){8-9}\cmidrule(lr){10-11}\cmidrule(lr){12-13}\cmidrule(lr){14-15}
    
    ~ & AbsRel & $\delta_1$ & AbsRel & $\delta_1$ & AbsRel & $\delta_1$ & AbsRel & $\delta_1$ & AbsRel & $\delta_1$ & AbsRel & $\delta_1$ & AbsRel & $\delta_1$ \\
    
    \midrule

    BlendedMVS~\cite{blendedmvs} & \emph{0.089} & 0.918 & 0.068 & 0.958 & \emph{0.556} & 0.689 & \emph{0.305} & \emph{0.731} & 0.148 & 0.845 & 0.092 & 0.921 & \emph{0.210} & \emph{0.844} \\
    
    DIML~\cite{diml} & 0.099 & 0.907 & \underline{0.055} & \emph{0.969} & 0.573 & 0.722 & 0.381 & 0.657 & \underline{0.142} & \underline{0.859} & 0.107 & 0.908 & 0.226 & 0.837 \\

    HRWSI~\cite{hrwsi} & 0.095 & 0.917 & 0.062 & 0.966 & \underline{0.502} & \underline{0.731} & \underline{0.270} & \underline{0.750} & 0.186 & 0.775 & \underline{0.087} & \underline{0.935} & \underline{0.200} & \underline{0.846} \\

    IRS~\cite{irs} & 0.105 & 0.892 & \emph{0.057} & \underline{0.970} & 0.568 & 0.714 & 0.328 & 0.691 & 0.143 & 0.845 & 0.088 & 0.926 & 0.215 & 0.840 \\
    
    MegaDepth~\cite{megadepth} & 0.217 & 0.741 & 0.071 & 0.953 & 0.632 & 0.660 & 0.479 & 0.566 & \emph{0.142} & \emph{0.852} & 0.104 & 0.910 & 0.274 & 0.780 \\

    TartanAir~\cite{tartanair} & \underline{0.088} & \underline{0.920} & 0.061 & 0.964 & 0.602 & \emph{0.723} & 0.332 & 0.690 & 0.160 & 0.818 & \emph{0.088} & \emph{0.928} & 0.222 & 0.841 \\

    \midrule

    All labeled data & \textbf{0.085} & \textbf{0.934} & \textbf{0.053} & \textbf{0.971} & \textbf{0.492} & \textbf{0.748} & \textbf{0.245} & \textbf{0.771} & \textbf{0.134} & \textbf{0.874} & \textbf{0.070} & \textbf{0.945} & \textbf{0.180} & \textbf{0.874} \\
    
    \bottomrule
    \end{tabular}
    \caption{Examine the zero-shot transferring performance of \emph{each labeled training set} (left) to six unseen datasets (top). \textbf{Better performance:} AbsRel $\downarrow$ , $\delta_1$ $\uparrow$. We highlight the \textbf{best}, \underline{second}, and \emph{third best} results for each test dataset in \textbf{bold}, \underline{underline}, and \emph{italic}, respectively.}
    \label{tab:each_trainset}
\end{table*}

\begin{table}[t]
\small
    \centering
    \setlength\tabcolsep{1.8mm}
    \begin{tabular}{rrcc}
    \toprule
    Method & Encoder & mIoU (s.s.) & m.s. \\

    \midrule

    Segmenter~\cite{segmenter} & ViT-L~\cite{vit} & - & 82.2 \\
    
    SegFormer~\cite{segformer} & MiT-B5~\cite{segformer} & 82.4 & 84.0 \\
    
    Mask2Former~\cite{mask2former} & Swin-L~\cite{swin} & 83.3 & 84.3 \\

    OneFormer~\cite{oneformer} & Swin-L~\cite{swin} & 83.0 & 84.4 \\
    
    OneFormer~\cite{oneformer} & ConvNeXt-XL~\cite{convnext} & 83.6 & 84.6 \\

    DDP \cite{ddp} & ConvNeXt-L~\cite{convnext} & 83.2 & 83.9 \\
    
    \midrule
    
    Ours & ViT-L~\cite{vit} & \textbf{84.8} & \textbf{86.2} \\
    
    \bottomrule
    \end{tabular}
    \caption{Transferring our MDE pre-trained encoder to \textbf{Cityscapes} for semantic segmentation. We \textbf{\emph{do not}} use Mapillary~\cite{mapillary} for pre-training. s.s./m.s.: single-/multi-scale evaluation.}
    \label{tab:cityscapes}
\end{table}

\subsection{Fine-tuned to Semantic Segmentation\label{sec:finetune_seg}}

In our method, we design our MDE model to inherit the rich semantic priors from a pre-trained encoder via a simple feature alignment constraint. Here, we examine the semantic capability of our MDE encoder. Specifically, we fine-tune our MDE encoder to downstream semantic segmentation datasets. As exhibited in Table~\ref{tab:cityscapes} of the Cityscapes dataset~\cite{cityscapes}, our encoder from large-scale MDE training (86.2 mIoU) is superior to existing encoders from large-scale ImageNet-21K pre-training, \eg, Swin-L~\cite{swin} (84.3) and ConvNeXt-XL~\cite{convnext} (84.6). Similar observations hold on the ADE20K dataset~\cite{ade20k} in Table~\ref{tab:ade20k}. We improve the previous best result from 58.3 $\rightarrow$ 59.4.

We hope to highlight that, witnessing the superiority of our pre-trained encoder on both monocular depth estimation and semantic segmentation tasks, we believe it has great potential to serve as a generic multi-task encoder for both middle-level and high-level visual perception systems.

\subsection{Ablation Studies\label{sec:ablation}}

Unless otherwise specified, we use the ViT-L encoder for our ablation studies here.

\vspace{2mm}
\noindent
\textbf{Zero-shot transferring of \emph{each} training dataset.}
In Table~\ref{tab:each_trainset}, we provide the zero-shot transferring performance of \emph{each} training dataset, which means that we train a relative MDE model on \emph{one} training set and evaluate it on the six unseen datasets. With these results, we hope to offer more insights for future works that similarly aim to build a general monocular depth estimation system. Among the six training datasets, HRWSI~\cite{hrwsi} fuels our model with the strongest generalization ability, even though it only contains 20K images. This indicates the data diversity counts a lot, which is well aligned with our motivation to utilize unlabeled images. Some labeled datasets may not perform very well, \eg, MegaDepth~\cite{megadepth}, however, it has its own preferences that are not reflected in these six test datasets. For example, we find models trained with MegaDepth data are specialized at estimating the distance of ultra-remote buildings (Figure~\ref{fig:teaser}), which will be very beneficial for aerial vehicles.

\vspace{2mm}
\noindent
\textbf{Effectiveness of 1) challenging the student model when learning unlabeled images, and 2) semantic constraint.} As shown in Table~\ref{tab:core_ablation}, simply adding unlabeled images with pseudo labels does not necessarily bring gains to our model, since the labeled images are already sufficient. However, with strong perturbations ($\mathcal{S}$) applied to unlabeled images during re-training, the student model is challenged to seek additional visual knowledge and learn more robust representations. Consequently, the large-scale unlabeled images enhance the model generalization ability significantly.

Moreover, with our used semantic constraint $\mathcal{L}_{feat}$, the power of unlabeled images can be further amplified for the depth estimation task. More importantly, as emphasized in Section~\ref{sec:finetune_seg}, this auxiliary constraint also enables our trained encoder to serve as a key component in a multi-task visual system for both middle-level and high-level perception.

\begin{table}[t]
\small
    \centering
    \setlength\tabcolsep{4mm}
    \begin{tabular}{rrc}
    \toprule
    Method & Encoder & mIoU \\

    \midrule

    Segmenter~\cite{segmenter} & ViT-L~\cite{vit} & 51.8 \\
    
    SegFormer~\cite{segformer} & MiT-B5~\cite{segformer} & 51.0 \\
    
    Mask2Former~\cite{mask2former} & Swin-L~\cite{swin} & 56.4 \\
    
    UperNet~\cite{upernet} & BEiT-L~\cite{beit} & 56.3 \\

    ViT-Adapter \cite{vit_adapter} & BEiT-L~\cite{beit} & 58.3 \\

    OneFormer~\cite{oneformer} & Swin-L~\cite{swin} & 57.4 \\

    OneFormer~\cite{oneformer} & ConNeXt-XL~\cite{convnext} & 57.4 \\
    
    \midrule
    
    Ours & ViT-L~\cite{vit} & \textbf{59.4} \\
    
    \bottomrule
    \end{tabular}
    \caption{Transferring our MDE encoder to \textbf{ADE20K} for semantic segmentation. We use Mask2Former as our segmentation model.}
    \label{tab:ade20k}
\end{table}

\vspace{2mm}
\noindent
\textbf{Comparison with MiDaS trained encoder in downstream tasks.}
Our Depth Anything model has exhibited stronger zero-shot capability than MiDaS~\cite{midas, midasv31}.
Here, we further compare our trained encoder with MiDaS v3.1~\cite{midasv31} trained encoder in terms of the downstream fine-tuning performance. As demonstrated in Table~\ref{tab:midas_downstream}, on both the downstream depth estimation task and semantic segmentation task, our produced encoder outperforms the MiDaS encoder remarkably, \eg, 0.951 \emph{vs.} 0.984 in the $\delta_1$ metric on NYUv2, and 52.4 \emph{vs.} 59.4 in the mIoU metric on ADE20K.

\begin{figure}[t]
    \centering
    \includegraphics[width=\linewidth]{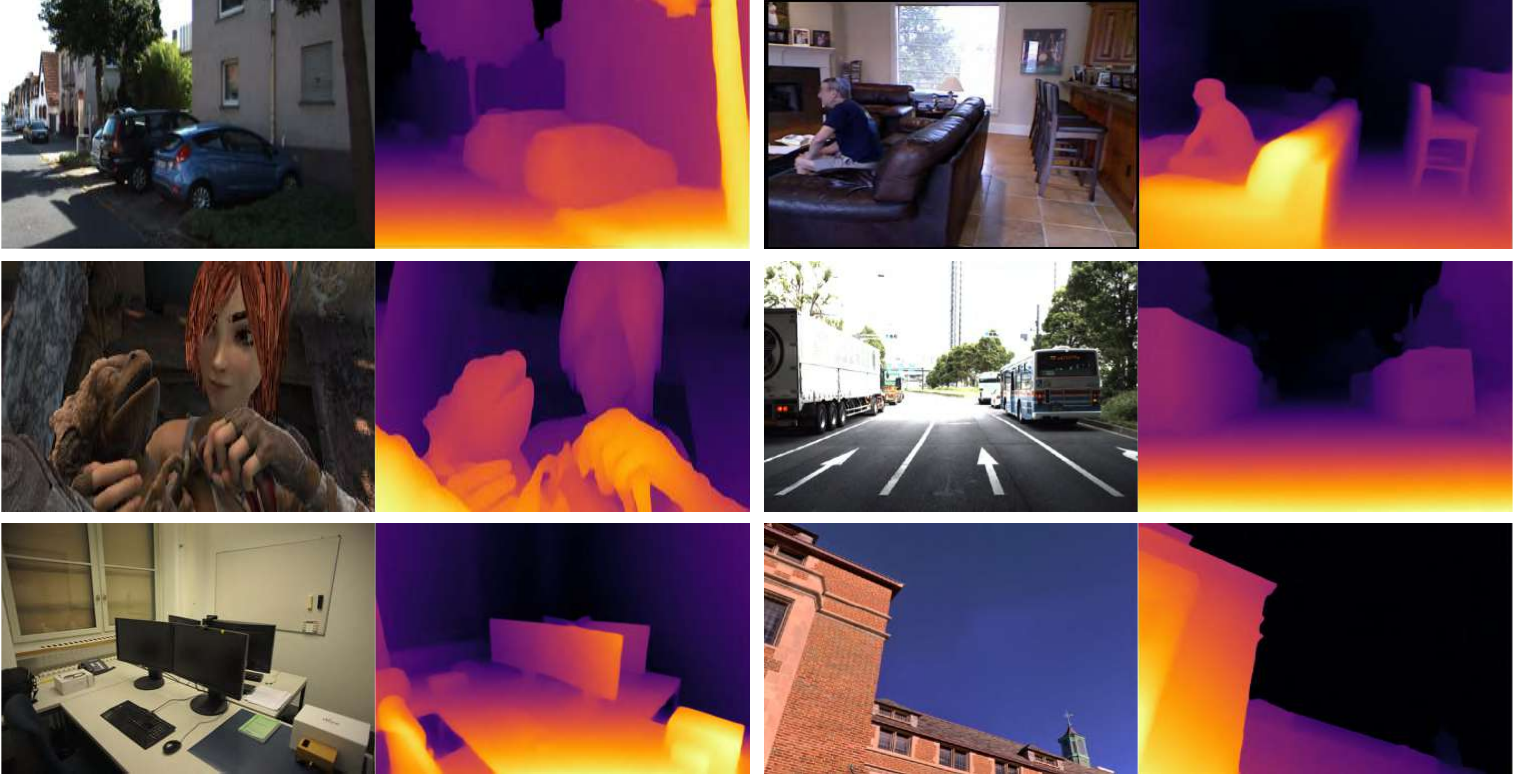}
    \caption{Qualitative results on six unseen datasets.}
    \label{fig:zeroshot_vis}
\end{figure}

\begin{table}[t]
\small
    \centering
    \setlength\tabcolsep{1.15mm}
    \begin{tabular}{cccccccccc}
    \toprule
    $\mathcal{L}_l$ & $\mathcal{L}_u$ & $\mathcal{S}$ & $\mathcal{L}_{feat}$ & KI & NY & SI & DD & ET & DI \\
    \midrule
    
    \cmark & & & & 0.085 & 0.053 & 0.492 & 0.245 & 0.134 & 0.070 \\
    
    \cmark & \cmark & & & 0.085 & 0.054 & 0.481 & 0.242 & 0.138 & 0.073 \\
    
    \cmark & \cmark & \cmark & & 0.081 & 0.048 & 0.469 & 0.235 & 0.134 & 0.068 \\
    
    \cmark & \cmark & \cmark & \cmark & \textbf{0.076} & \textbf{0.043} & \textbf{0.458} & \textbf{0.230} & \textbf{0.127} & \textbf{0.066} \\
    
    \bottomrule
    \end{tabular}
    \caption{Ablation studies of: 1) challenging the student with strong perturbations ($\mathcal{S}$) when learning unlabeled images, and 2) semantic constraint ($\mathcal{L}_{feat}$). Limited by space, we only report the AbsRel ($\downarrow$) metric, and shorten the dataset name with its first two letters.}
    \label{tab:core_ablation}
\end{table}

\begin{table}[t]
\small
    \centering
    \setlength\tabcolsep{1mm}
    \begin{tabular}{lcccccc}
    \toprule
    
    \multirow{2}{*}{Method} & \multicolumn{2}{c}{NYUv2} & \multicolumn{2}{c}{KITTI} & Cityscapes & ADE20K \\
    
    \cmidrule(lr){2-3}\cmidrule(lr){4-5}\cmidrule(lr){6-6}\cmidrule(lr){7-7}
    
    ~ & AbsRel & $\delta_1$ & AbsRel & $\delta_1$ & mIoU & mIoU \\
    
    \midrule

    MiDaS & 0.077 & 0.951 & 0.054 & 0.971 & 82.1 & 52.4 \\

    Ours & \textbf{0.056} & \textbf{0.984} & \textbf{0.046} & \textbf{0.982} &  \textbf{84.8} & \textbf{59.4} \\
    
    \bottomrule
    \end{tabular}
    \caption{Comparison between our trained encoder and MiDaS~\cite{midasv31} trained encoder in terms of downstream fine-tuning performance. \textbf{Better performance:} AbsRel $\downarrow$ , $\delta_1 \uparrow$ , mIoU $\uparrow$ .}
    \label{tab:midas_downstream}
\end{table}

\vspace{2mm}
\noindent
\textbf{Comparison with DINOv2 in downstream tasks.}
We have demonstrated the superiority of our trained encoder when fine-tuned to downstream tasks. Since our finally produced encoder (from large-scale MDE training) is fine-tuned from DINOv2~\cite{dinov2}, we compare our encoder with the original DINOv2 encoder in Table~\ref{tab:dinov2_downstream}. It can be observed that our encoder performs better than the original DINOv2 encoder in both the downstream metric depth estimation task and semantic segmentation task. 
Although the DINOv2 weight has provided a very strong initialization, our large-scale and high-quality MDE training can further enhance it impressively in downstream transferring performance.

\begin{figure}[t]
    \centering
    \includegraphics[width=\linewidth]{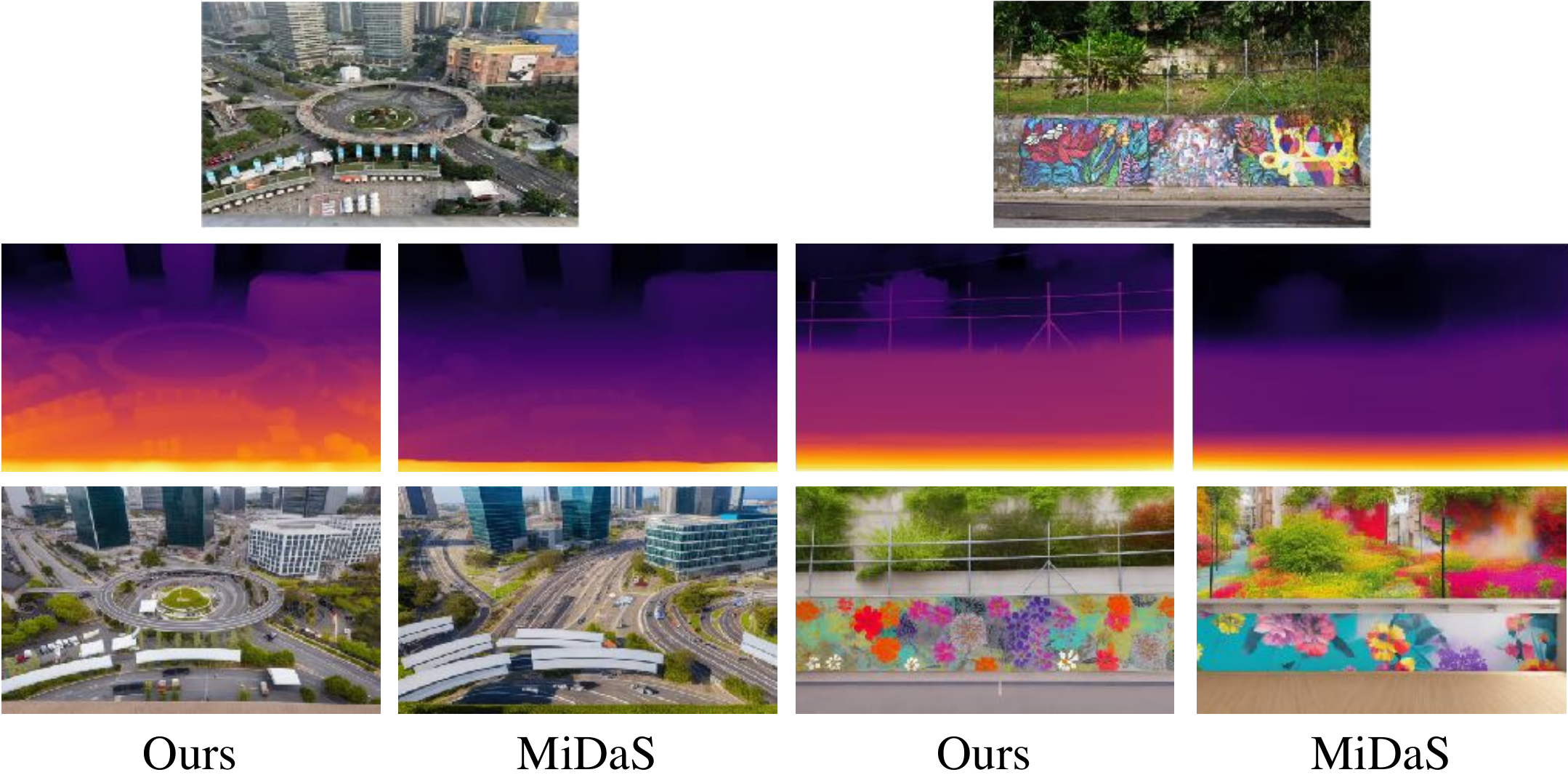}
    \caption{We compare our depth prediction with MiDaS. Meantime, we use ControlNet to synthesize new images from the depth map.}
    \label{fig:synthesis}
\end{figure}

\begin{table}[t]
\small
    \centering
    \setlength\tabcolsep{1.8mm}
    \begin{tabular}{lccccc}
    \toprule
    \multirow{2}{*}{Encoder} & \multicolumn{2}{c}{NYUv2} & \multicolumn{2}{c}{KITTI} & ADE20K \\

    \cmidrule(lr){2-3}\cmidrule(lr){4-5}\cmidrule(lr){6-6}
    
    ~ & AbsRel ($\downarrow$) & $\delta_1$ ($\uparrow$) & AbsRel & $\delta_1$ & mIoU ($\uparrow$) \\

    \midrule
    
    DINOv2 & 0.066 & 0.973 & 0.058 & 0.971 & 58.8 \\
    
    Ours & \textbf{0.056} & \textbf{0.984} & \textbf{0.046} & \textbf{0.982} & \textbf{59.4} \\
    
    \bottomrule
    \end{tabular}
    \caption{Comparison between the original DINOv2 and our produced encoder in terms of downstream fine-tuning performance.}
    \label{tab:dinov2_downstream}
\end{table}

\subsection{Qualitative Results\label{sec:qualitative}}

We visualize our model predictions on the six unseen datasets in Figure~\ref{fig:zeroshot_vis}. Our model is robust to test images from various domains. In addition, we compare our model with MiDaS in Figure~\ref{fig:synthesis}. We also attempt to synthesis new images conditioned on the predicted depth maps with ControlNet~\cite{controlnet}. Our model produces more accurate depth estimation than MiDaS, as well as better synthesis results. For more accurate synthesis, we re-trained a better depth-conditioned ControlNet based on our Depth Anything, aiming to provide better control signals for image synthesis and video editing. Please refer to our project page for more qualitative results on video editing~\cite{magicedit} with our Depth Anything.
\section{Conclusion}

In this work, we present Depth Anything, a highly practical solution to robust monocular depth estimation. Different from prior arts, we especially highlight the value of cheap and diverse unlabeled images. We design two simple yet highly effective strategies to fully exploit their value: 1) posing a more challenging optimization target when learning unlabeled images, and 2) preserving rich semantic priors from pre-trained models. As a result, our Depth Anything model exhibits excellent zero-shot depth estimation ability, and also serves as a promising initialization for downstream metric depth estimation and semantic segmentation tasks.

\vspace{3mm}
\noindent
\textbf{Acknowledgement.} This work is supported by the National Natural Science Foundation of China (No. 62201484), HKU Startup Fund, and HKU Seed Fund for Basic Research.
\clearpage
\setcounter{page}{9}
\maketitlesupplementary

\section{More Implementation Details}

We resize the shorter side of all images to 518 and keep the original aspect ratio. All images are cropped to 518$\times$518 during training. During inference, we do not crop images and only ensure both sides are multipliers of 14, since the pre-defined patch size of DINOv2 encoders~\cite{dinov2} is 14. Evaluation is performed at the original resolution by interpolating the prediction. Following MiDaS~\cite{midas, midasv31}, in zero-shot evaluation, the scale and shift of our prediction are manually aligned with the ground truth.

When fine-tuning our pre-trained encoder to metric depth estimation, we adopt the ZoeDepth codebase~\cite{zoedepth}. We merely replace the original MiDaS-based encoder with our stronger Depth Anything encoder, with a few hyper-parameters modified. Concretely, the training resolution is 392$\times$518 on NYUv2~\cite{nyu} and 384$\times$768 on KITTI~\cite{kitti} to match the patch size of our encoder. The encoder learning rate is set as 1/50 of the learning rate of the randomly initialized decoder, which is much smaller than the 1/10 adopted for MiDaS encoder, due to our strong initialization. The batch size is 16 and the model is trained for 5 epochs.

When fine-tuning our pre-trained encoder to semantic segmentation, we use the MMSegmentation codebase~\cite{mmseg}. The training resolution is set as 896$\times$896 on both ADE20K~\cite{ade20k} and Cityscapes~\cite{cityscapes}. The encoder learning rate is set as 3e-6 and the decoder learning rate is 10$\times$ larger. We use Mask2Former~\cite{mask2former} as our semantic segmentation model. The model is trained for 160K iterations on ADE20K and 80K iterations on Cityscapes both with batch size 16, without any COCO~\cite{coco} or Mapillary~\cite{mapillary} pre-training. Other training configurations are the same as the original codebase.

\section{More Ablation Studies}

All ablation studies here are conducted on the ViT-S model.

\vspace{1mm}
\noindent
\textbf{The necessity of tolerance margin for feature alignment.} 
As shown in Table~\ref{tab:tolerance_margin}, the gap between the tolerance margin of 1.00 and 0.85 or 0.70 clearly demonstrates the necessity of this design (mean AbsRel: 0.188 \emph{vs.} 0.175).

\vspace{1mm}
\noindent
\textbf{Applying feature alignment to labeled data.} Previously, we enforce the feature alignment loss $\mathcal{L}_{feat}$ on unlabeled data. Indeed, it is technically feasible to also apply this constraint to labeled data. In Table~\ref{tab:featloss}, apart from applying $\mathcal{L}_{feat}$ on unlabeled data, we explore to apply it to labeled data. We find that adding this auxiliary optimization target to labeled data is not beneficial to our baseline that does not involve any feature alignment (their mean AbsRel values are almost the same: 0.180 \emph{vs.} 0.179). We conjecture that this is because the labeled data has relatively higher-quality depth annotations. The involvement of semantic loss may interfere with the learning of these informative manual labels. In comparison, our pseudo labels are noisier and less informative. Therefore, introducing the auxiliary constraint to unlabeled data can combat the noise in pseudo depth labels, as well as arm our model with semantic capability.

\begin{table}[t]
\small
    \centering
    \setlength\tabcolsep{1.15mm}
    \begin{tabular}{lccccccc}
    \toprule
    $\alpha$ & KITTI & NYU & Sintel & DDAD & ETH3D & DIODE & \textbf{Mean} \\
    \midrule
    
    1.00 & 0.085 & 0.055 & 0.523 & 0.250 & 0.134 & 0.079 & 0.188 \\

    0.85 & 0.080 & \textbf{0.053} & \textbf{0.464} & \textbf{0.247} &  \textbf{0.127} &  \textbf{0.076} & \textbf{0.175} \\

    0.70 & \textbf{0.079} & 0.054 & 0.482 & 0.248 & 0.127 & 0.077 & 0.178 \\
    
    \bottomrule
    \end{tabular}
    \caption{Ablation studies on different values of the tolerance margin $\alpha$ for the feature alignment loss $\mathcal{L}_{feat}$. Limited by space, we only report the AbsRel ($\downarrow$) metric here.}
    \label{tab:tolerance_margin}
\end{table}

\begin{table}[t]
\small
    \centering
    \setlength\tabcolsep{1.05mm}
    \begin{tabular}{ccccccccc}
    \toprule
    
    \multicolumn{2}{c}{$\mathcal{L}_{feat}$} & \multicolumn{6}{c}{Unseen datasets (AbsRel $\downarrow$)} & \multirow{2}{*}{\textbf{Mean}} \\
    
    \cmidrule(lr){1-2}\cmidrule(lr){3-8}
    
    U & L & KITTI & NYU & Sintel & DDAD & ETH3D & DIODE &  \\
    
    \midrule

    & & 0.083 & 0.055 & 0.478 & 0.249 & 0.133 & 0.080 & 0.180 \\
    
    \cmark & & \textbf{0.080} & \textbf{0.053} & \textbf{0.464} & \textbf{0.247} & \textbf{0.127} & \textbf{0.076} & \textbf{0.175} \\

    & \cmark & 0.084 & 0.054 & 0.472 & 0.252 & 0.133 & 0.081 & 0.179 \\
    
    \bottomrule
    \end{tabular}
    \caption{Ablation studies of applying our feature alignment loss $\mathcal{L}_{feat}$ to unlabeled data (\textbf{U}) or labeled data (\textbf{L}).}
    \vspace{-1mm}
    \label{tab:featloss}
\end{table}

\section{Limitations and Future Works}

Currently, the largest model size is only constrained to ViT-Large~\cite{vit}. Therefore, in the future, we plan to further scale up the model size from ViT-Large to ViT-Giant, which is also well pre-trained by DINOv2~\cite{dinov2}. We can train a more powerful teacher model with the larger model, producing more accurate pseudo labels for smaller models to learn, \eg, ViT-L and ViT-B. Furthermore, to facilitate real-world applications, we believe the widely adopted 512$\times$512 training resolution is not enough. We plan to re-train our model on a larger resolution of 700+ or even 1000+.

\section{More Qualitative Results}

\begin{figure*}[t]
    \centering
    \includegraphics[width=\linewidth]{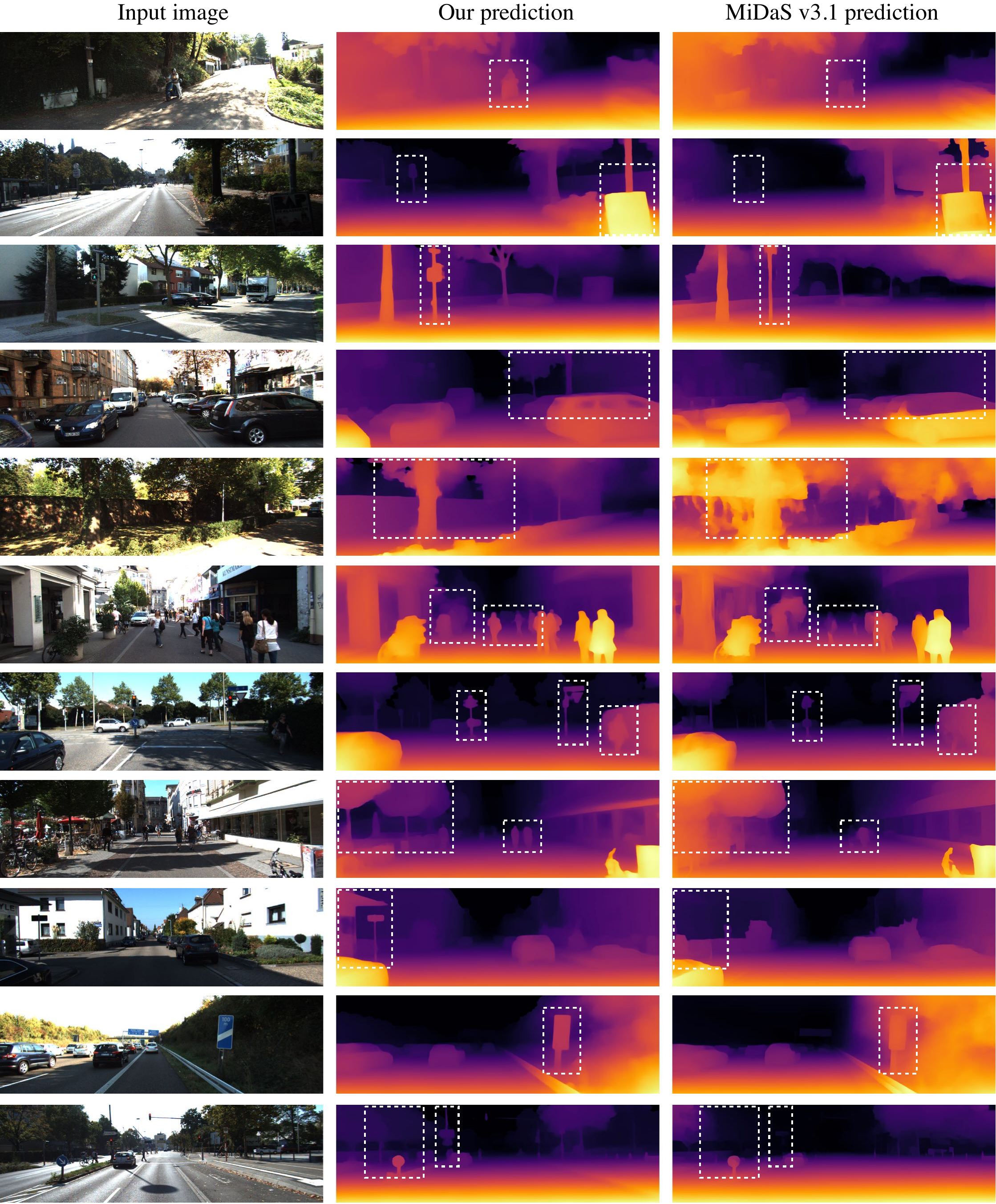}
    \caption{Qualitative results on KITTI. Due to the extremely sparse ground truth which is hard to visualize, we here compare our prediction with the most advanced MiDaS v3.1~\cite{midasv31} prediction. The brighter color denotes the closer distance.}
    \label{fig:vis_kitti}
\end{figure*}

\begin{figure*}[t]
    \centering
    \includegraphics[width=\linewidth]{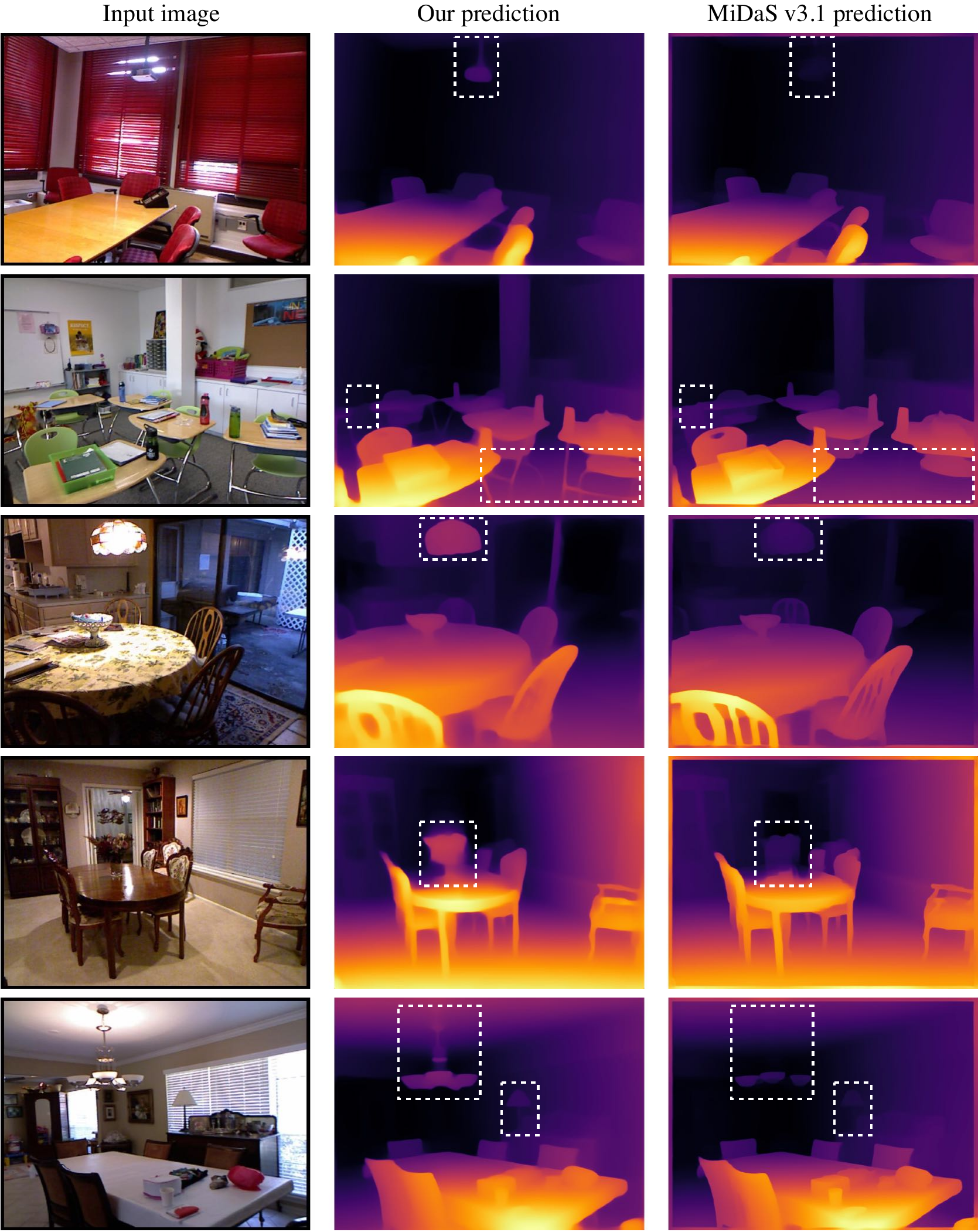}
    \caption{Qualitative results on NYUv2. It is worth noting that MiDaS~\cite{midasv31} uses NYUv2 training data (\emph{not zero-shot}), while we do not.}
    \label{fig:vis_nyu}
\end{figure*}

\begin{figure*}[t]
    \centering
    \includegraphics[width=\linewidth]{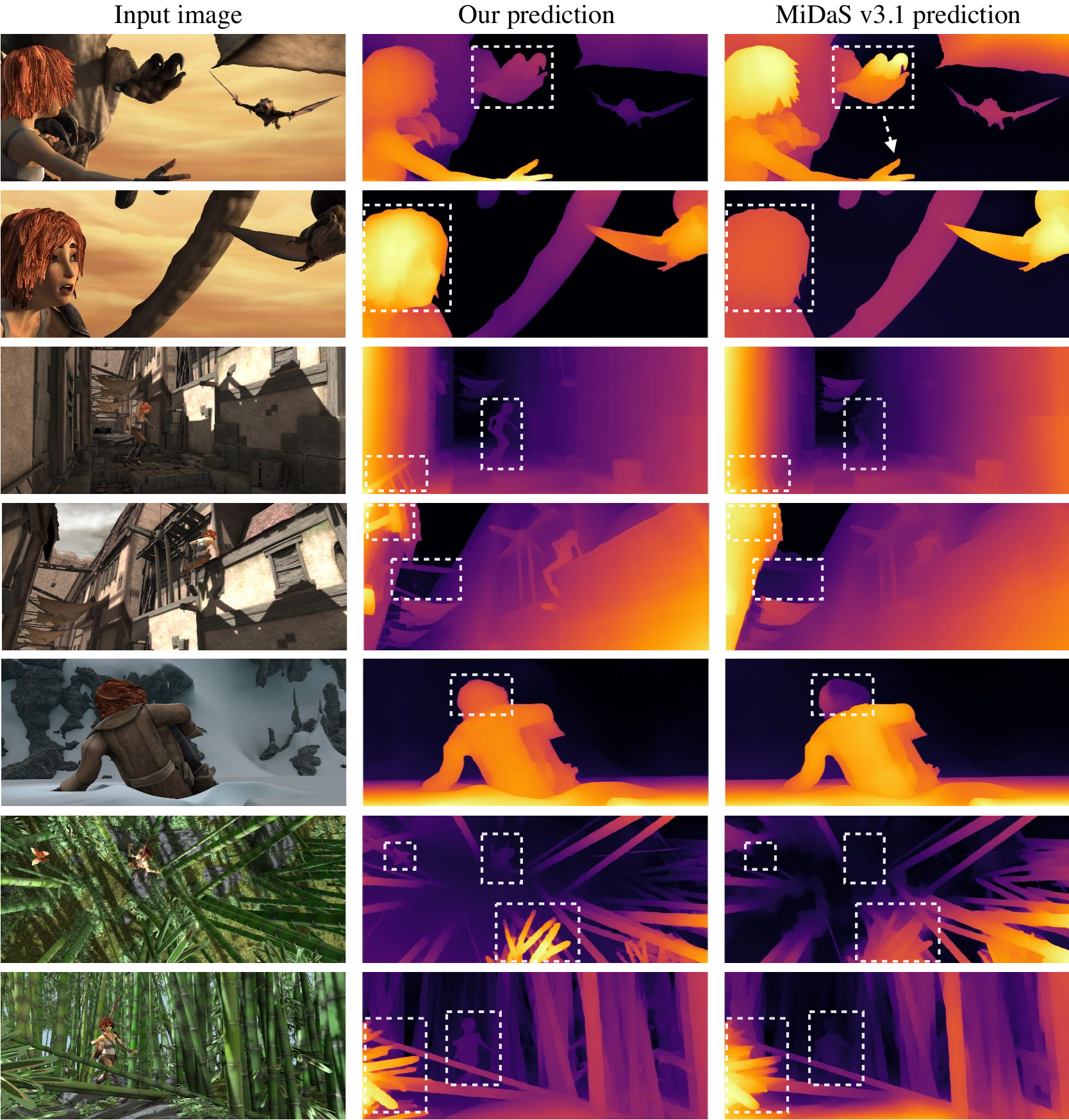}
    \caption{Qualitative results on Sintel.}
    \label{fig:vis_sintel}
\end{figure*}

\begin{figure*}[t]
    \centering
    \includegraphics[width=\linewidth]{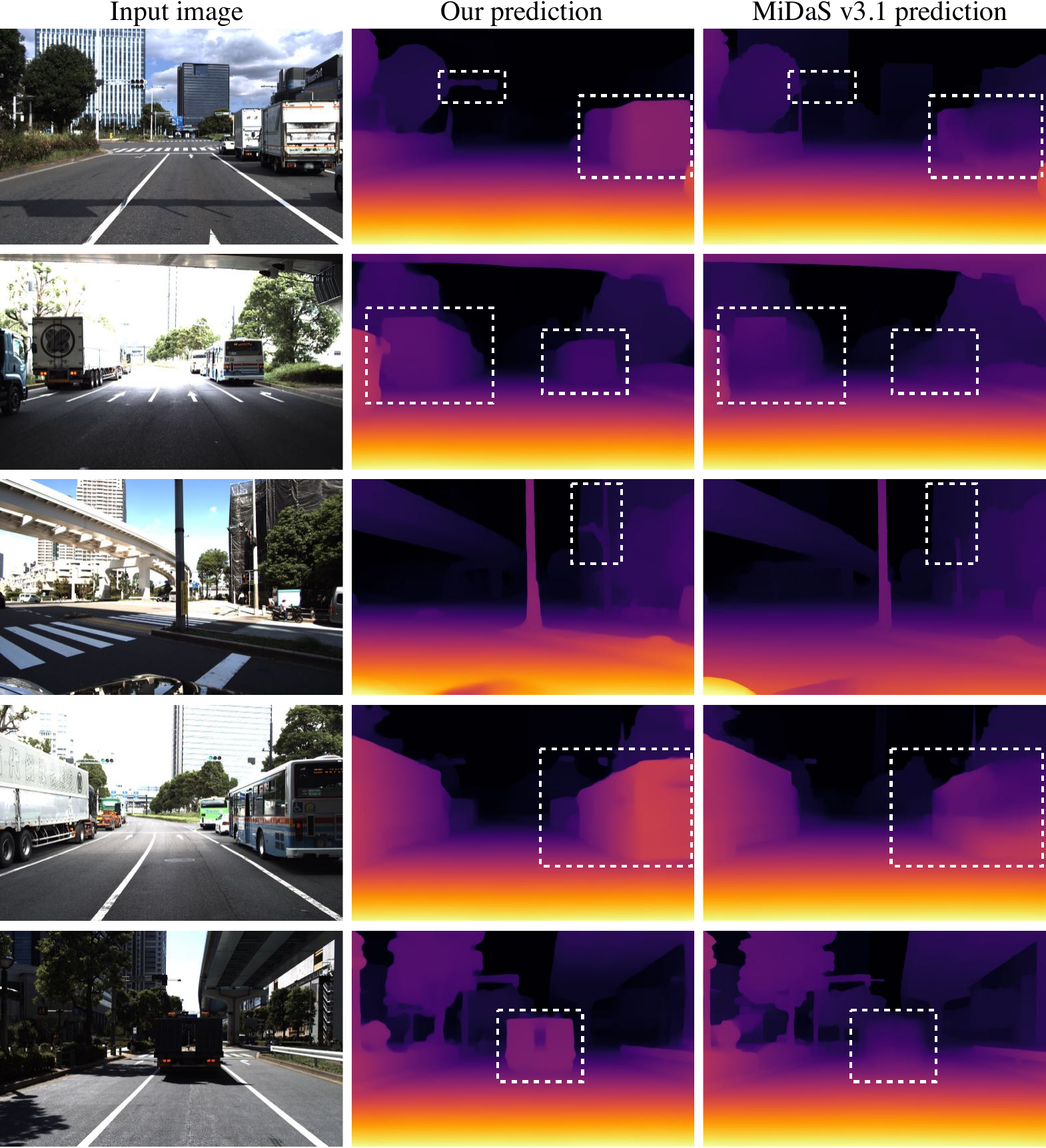}
    \caption{Qualitative results on DDAD.}
    \label{fig:vis_ddad}
\end{figure*}

\begin{figure*}[t]
    \centering
    \vspace{-3mm}
    \includegraphics[width=\linewidth]{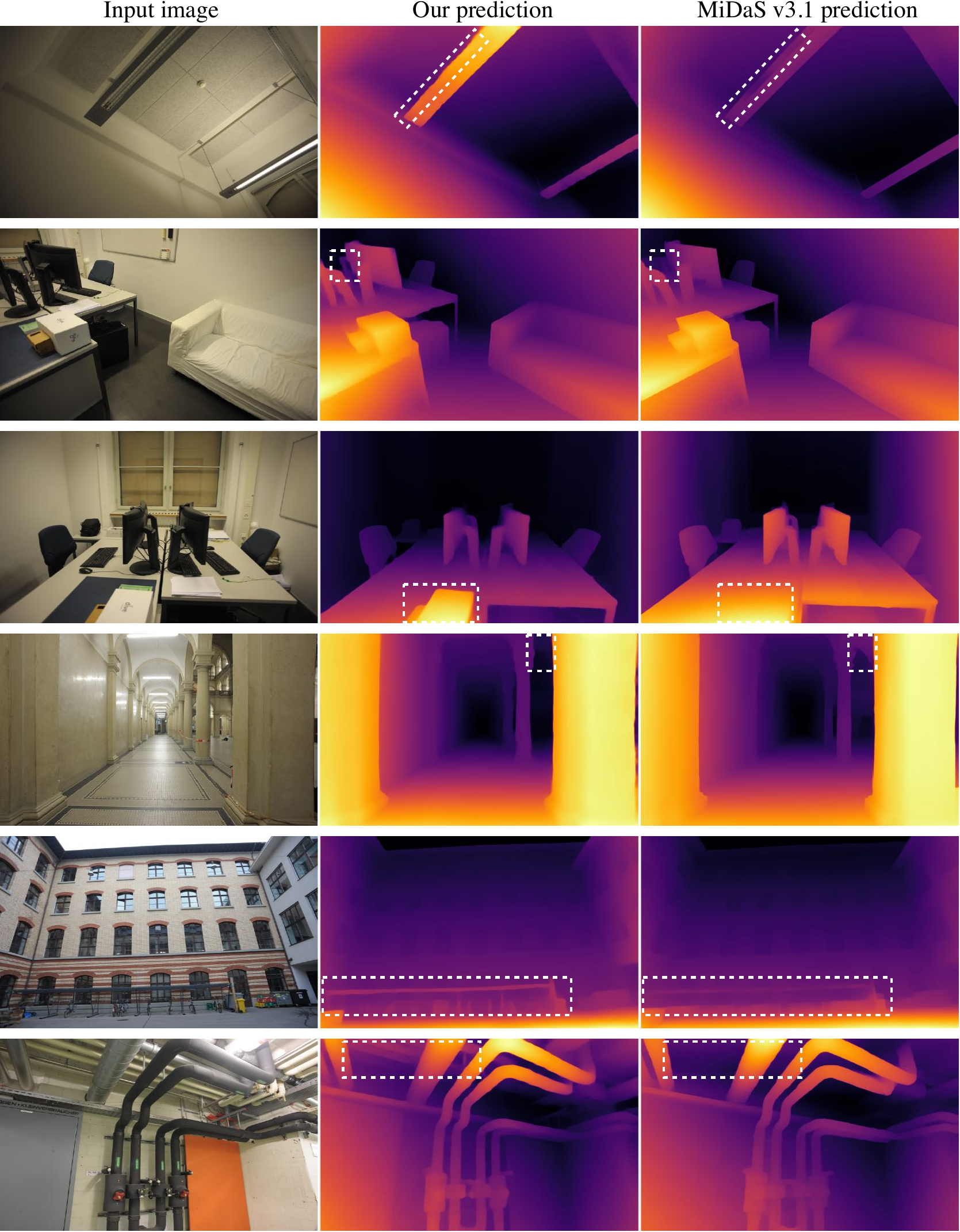}
    \caption{Qualitative results on ETH3D.}
    \label{fig:vis_eth3d}
\end{figure*}

\begin{figure*}[t]
    \centering
    \includegraphics[width=\linewidth]{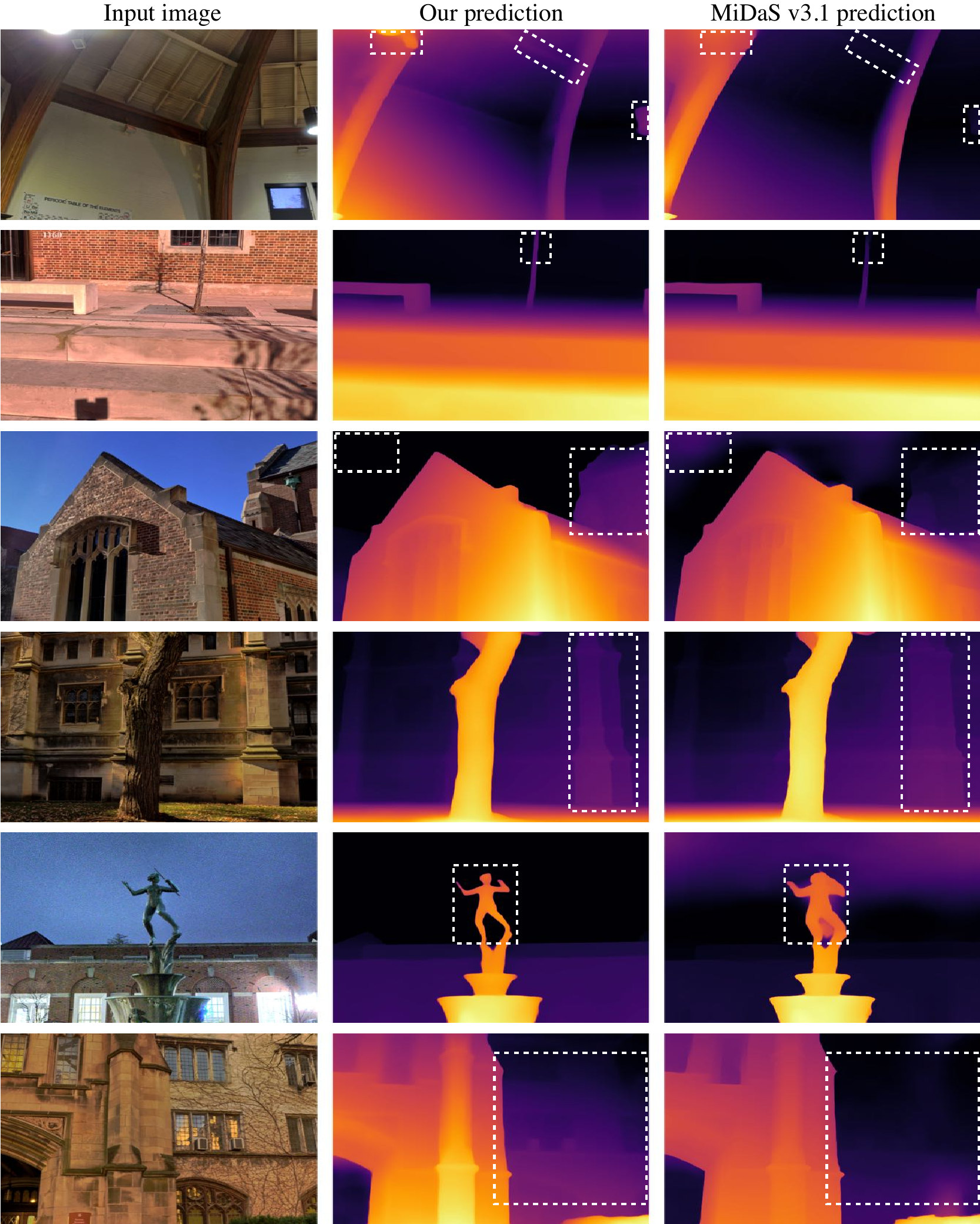}
    \caption{Qualitative results on DIODE.}
    \label{fig:vis_diode}
\end{figure*}

Please refer to the following pages for comprehensive qualitative results on six unseen test sets (Figure~\ref{fig:vis_kitti} for KITTI~\cite{kitti}, Figure~\ref{fig:vis_nyu} for NYUv2~\cite{nyu}, Figure~\ref{fig:vis_sintel} for Sintel~\cite{sintel}, Figure~\ref{fig:vis_ddad} for DDAD~\cite{ddad}, Figure~\ref{fig:vis_eth3d} for ETH3D~\cite{eth3d}, and Figure~\ref{fig:vis_diode} for DIODE~\cite{diode}). We compare our model with the strongest MiDaS model~\cite{midasv31}, \ie, DPT-BEiT$_{\textrm{L-512}}$. Our model exhibits higher depth estimation accuracy and stronger robustness. Please refer to our project page for more visualizations.

\clearpage
\clearpage
\newpage
{
    \small
    \bibliographystyle{ieeenat_fullname}
    \bibliography{main}
}

\end{document}